\documentclass[11pt,a4paper]{article}
\usepackage[hyperref]{ranlp2025}
\usepackage{times}
\usepackage{latexsym}

\usepackage{enumitem}
\usepackage{amsmath}
\usepackage{graphicx}
\usepackage{booktabs}

\usepackage{microtype}

\aclfinalcopy 
\def\aclpaperid{53} 

\title{Measuring How (Not Just Whether) VLMs Build Common Ground}

\author{
    Saki Imai \qquad Mert İnan \qquad Anthony Sicilia \qquad Malihe Alikhani \\
    Northeastern University, Boston MA \\
    \texttt{\{imai.s, m.alikhani\}@northeastern.edu}
}

\date{}

\begin{document}
\maketitle
\begin{abstract}
Large vision language models (VLMs) increasingly claim reasoning skills, yet current benchmarks evaluate them in single-turn or question answering settings. However, grounding is an interactive process in which people gradually develop shared understanding through ongoing communication. We introduce a four-metric suite (\emph{grounding efficiency}, \emph{content alignment}, \emph{lexical adaptation}, and \emph{human-likeness}) to systematically evaluate VLM performance in interactive grounding contexts. We deploy the suite on 150 self-play sessions of interactive referential games between three proprietary VLMs and compare them with human dyads. All three models diverge from human patterns on at least three metrics, while GPT4o-mini is the closest overall. We find that (i) task success scores do not indicate successful grounding and (ii) high image-utterance alignment does not necessarily predict task success. Our metric suite and findings offer a framework for future research on VLM grounding.
\end{abstract} 

\begin{figure}[!h]
    \centering
    \includegraphics[width=\linewidth]{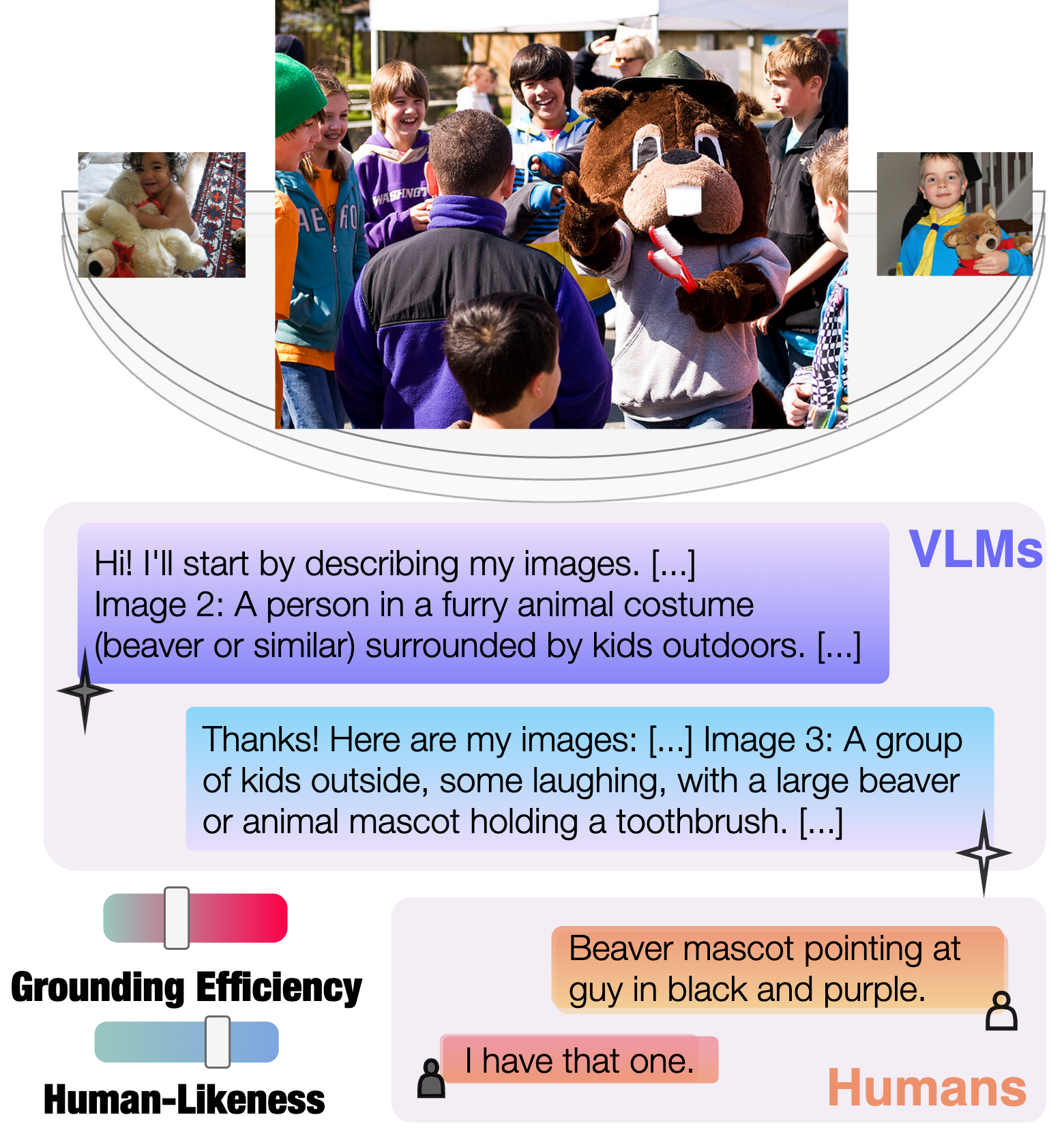}
    \caption{This figure shows our VLM evaluation suite using the Photobook Task to test the grounding capabilities and human-likeness of language models, in addition to two other metrics. Our benchmarking strategy clearly identifies the differences between VLM and human interactions.}
    \label{fig:fig1}
\end{figure}

\section{Introduction}
To build collaborative AI systems, it is not enough to produce locally correct answers; agents must \emph{establish common ground} efficiently through interaction, as humans do in situated dialogue \citep{clark1991grounding}.\footnote{Following \citet{clark1996using}, we use \emph{common ground} to mean the set of propositions that all interlocutors mutually believe, know that the others believe, and recognize as a basis for subsequent action.} Human partners achieve this via rapid lexical entrainment \cite{Brennan, krauss1964changes, brennan1996lexical, garrod1987saying}, and multi-level interactive alignment \citep{Pickering_Garrod_2004}, that yields shorter, more precise utterances over time. These are shown to both increase task success \cite{reitter-moore-2007-predicting} and reduce cognitive load.

Despite growing support for multi-turn interaction, contemporary training and evaluation pipelines for large language and vision language models (VLMs) still prioritize single-turn answer quality, through supervised fine‑tuning, RLHF \cite{ouyang2022training}, or DPO \citep{rafailov2023direct}. As a result, they neither measure nor reward the interactive skills that underpin grounding, such as reusing a partner’s words or pruning redundant detail once mutual understanding is achieved. Recent evidence also points to low communication efficiency and degraded multi-turn performance relative to single-turn settings \citep{hua2024talk, laban2025llms}.


In this paper, we operationalize grounding for multimodal dialogue and evaluate it directly. We introduce a task-agnostic evaluation metric suite that captures \textit{grounding efficiency}, \textit{content alignment}, \textit{lexical adaptation}, and \textit{human-likeness}—and instantiate it on the PhotoBook referential game \citep{haber-etal-2019-photobook}, which features five-round dialogues to identify shared images (Fig.~\ref{fig:fig1}; \S\ref{sec:metrics}). \footnote{\url{https://github.com/sakimai/vlm-grounding-benchmark}} 
While the original Photobook corpus studied the common ground formulation and referring expression generation using LSTM models, they did not (i) test with contemporary billion-scale VLMs, (ii) allow model-model self-play, or (iii) quantify distributional human-likeliness. We benchmark contemporary VLMs in model-model self-play and compare against human transcripts to ask:
\begin{enumerate}[noitemsep, nolistsep]
    \item How efficiently do VLM pairs reach common ground compared to humans? (grounding efficiency)
    \item Do VLMs describe the exact visual cues and is that predictive of task success? (content alignment)
    \item Do VLM pairs form human-like conceptual pacts, reusing each other's terms and pruning redundant detail over rounds? (lexical adaptation)
    \item To what extent do the grounding behaviors of VLM pairs resemble human dialogue patterns at the distributional level? (human-likeliness)
\end{enumerate}
\vspace{0.5em}

We show that VLM diverges from human baselines on $\geq 3$ metrics, where GPT4o-mini is closest overall (\S\ref{sec:results}). Notably, high image-utterance alignment does not guarantee task success—there is no correlation between CLIPScore and task outcomes (\S\ref{ssec:res_alignment}). Finally, we show that task success does not imply grounding. GPT4.1 often inflates its score by mirroring partner's preferences when ground-truth labels coincide (\S\ref{sec:case_study}).

\section{Related Work}
\paragraph{Common ground and lexical entrainment}
Many lines of research in cognitive science and linguistics have focused on modeling common ground establishment in human-human interactions. Studies demonstrated that conversational partners converge on concise, mutually understood referring expressions across successive turns~\citep{krauss1964changes, CLARK19861, brennan1996lexical, garrod1987saying}. The interactive alignment model \citep{Pickering_Garrod_2004} proposed that lexical, syntactic and discourse level alignments emerge and support higher level coordination and mutual understanding.

Computational models have attempted to replicate these behaviors in task oriented dialogue systems \citep{stoyanchev-stent-2009-lexical, devault2011incremental, visser2014model, ohashi2022adaptive}. However, much of this work has focused on text or spoken dialogue systems. Large VLMs have only recently been examined for lexical adaptation \citep{hua2024talk}. Our work builds on these insights by operationalizing common ground into four concrete metrics.

\paragraph{Visual reference games}
Reference games provide a controlled environment to study grounding processes. In these tasks, participants must identify shared referents, through dialogue \citep{krauss1964changes, CLARK19861, hawkins2017convention, monroe-etal-2017-colors}. These tasks have been adapted for computational modeling to evaluate alignment, reasoning, and visual understanding in interactive contexts \citep{he-etal-2017-learning, hawkins2017convention, hawkins-etal-2020-continual}. The PhotoBook dataset \citep{haber-etal-2019-photobook}, used in our study, extends this paradigm to three-round dialogues. This offers an environment to study common ground formation across time.

More recent work has explored reference games for model evaluation in both abstract and grounded domains \citep{ji-etal-2022-abstract, chalamalasetti-etal-2023-clembench, hakimov-etal-2025-using}. However, these approaches typically treat interaction as a means to an end, without probing how communicative strategies evolve. In contrast, we introduce a new set of metrics and visualizations to trace the evolution of grounding behaviors and lexical strategies across dialogue rounds.


\begin{figure*}
    \centering
    \includegraphics[width=0.95\linewidth]{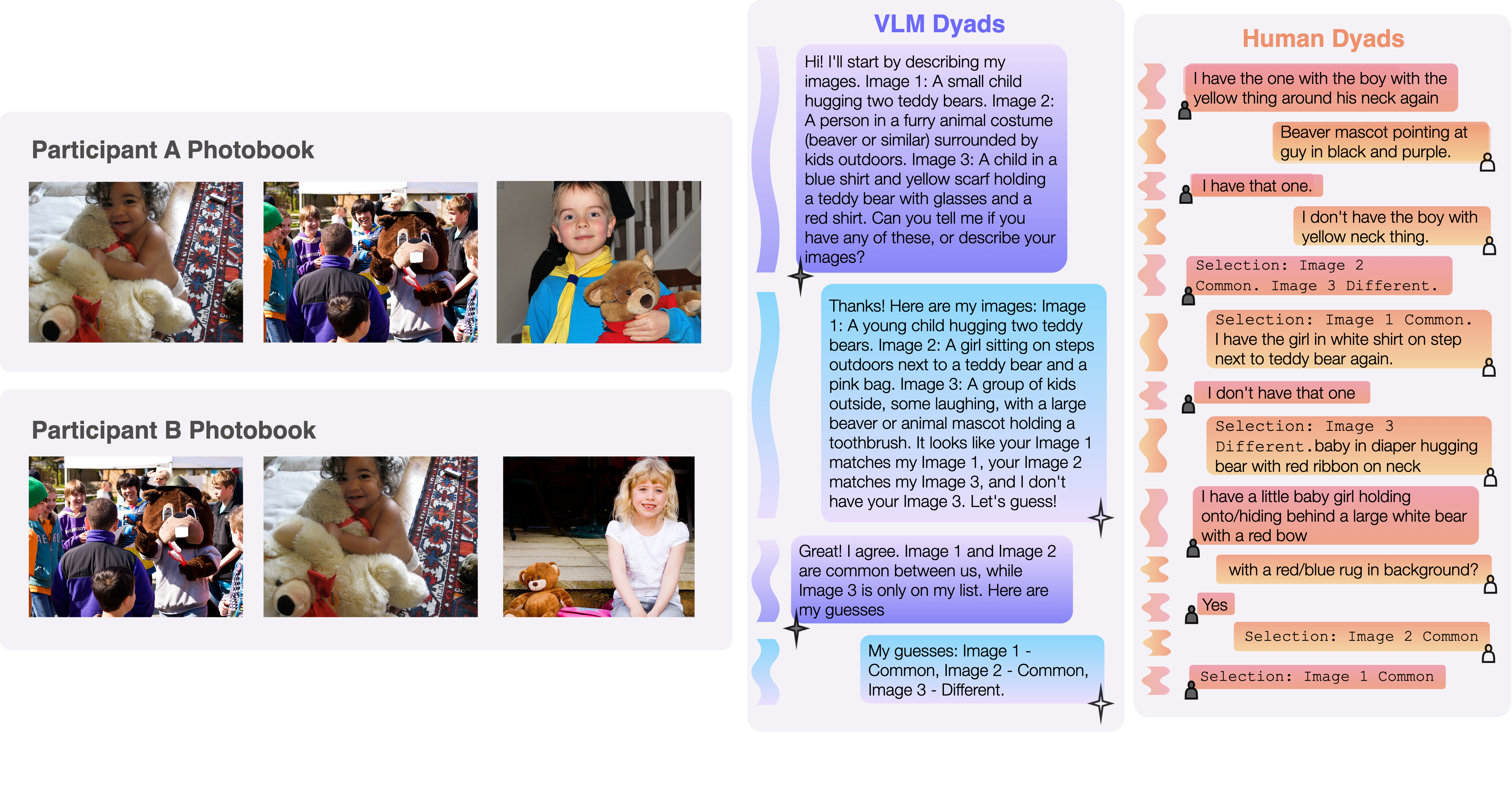}
    \vspace{-30pt}
    \caption{This figure shows a comparative example of human-human and VLM dyadic conversations on the same Photobook game. Human dialogues are more incremental, contain shorter utterances, and have more turns, while VLMs engage in long explanations for each image, with fewer turns, and with sycophancy. The images are from the MSCOCO dataset \cite{lin2014microsoft}.}
    \label{fig:photobook_example}
\vspace{-10pt}
\end{figure*}

\paragraph{VLM evaluation in multimodal interactions}
While large‑scale VLMs have achieved impressive zero‑shot accuracy on static benchmarks \cite{liu2023visual, achiam2023gpt, liu2024improved}, we still lack an understanding of how they behave in extended, collaborative interaction \cite{sicilia2022modeling}. Recent work suggests that VLMs fall short of interactive human behaviors. \citet{hua2024talk}, for example, show that even the most capable multimodal models struggle to adapt to their partner's word choices in-context, and tend to produce verbose but low-efficiency utterances. Our study simulates VLM-VLM dialogues in the PhotoBook game, comparing them to human interactions. This approach, unlike prior work using synthetic or asymmetrical dialogue, quantifies how VLMs build common ground, not just if they do.

\section{Grounding Evaluation Suite}
\label{sec:framework}
This section details the controlled setting we use to examine how VLMs build common ground in dialogue (see Figure~\ref{fig:photobook_example} for an example scenario).

\subsection{The PhotoBook Task}
\paragraph{Task.}
PhotoBook is a five round referential game in which two conversational partners must discover which of three images they share, and which are unique to each speaker \citep{haber-etal-2019-photobook}. At each round, they privately annotate every image as \textit{common} or \textit{different}. The images in the game extracted from the MS COCO Dataset \citep{lin2014microsoft} are deliberately structured to be visually similar to elicit non-trivial referring expressions. Moreover, the images appear exactly five times throughout a game, which allow us to study the collaborative referring expression generation and resolution \citep{CLARK19861}. 

\paragraph{Human corpus.}
The released dataset contains 2,506 human–human dialogues (164,615 utterances, 130,322 actions and spans a vocabulary of 11,805 unique tokens). This serves as an empirical upper bound baseline for grounding efficiency. Additional work on referring expression extracted 41,340 referring utterances and 16,525 chains from this dataset \citep{takmaz-etal-2020-refer}.

\subsection{VLM Self‑Play Protocol}
\label{ssec:dyads}
\paragraph{Models.}
We study three recent proprietary VLMs that differ in size and architecture: (i) GPT4.1 (ii) GPT4o-mini (iii) Claude3.5-Haiku.
Each VLM dyad initialized with default parameters plays the same set of 50 games, resulting in a total of 150 games. This setup enables a comparison of lexical strategies, task performance, and communicative behavior between VLMs and human.

\paragraph{Prompting and turn scheduling.}
We instantiate two agents that alternate turns until both submit non‑null guesses. Each turn must be a valid JSON object:
\begin{itemize}[noitemsep,leftmargin=*,nolistsep]
  \item \texttt{"message"}  the natural‑language utterance,  
  \item \texttt{"reference"}  either "Image $k$" ($k\in\{1,2,3\}$) or
        \texttt{null},
  \item \texttt{"guesses"}  \texttt{null} until a player is ready, otherwise a
        three‑letter array such as \texttt{"C","D","C"} where
        \texttt{"C"} $\equiv$ common, \texttt{"D"} $\equiv$ different.
\end{itemize}
The full prompt is in Appendix~\ref{sec:app_prompt}. 
A prompt engineered variant designed to prevent three recurrent failure cases, (1) prematurely revealing guesses, (2) comparing images one by one rather than as a set, (3) generating fillers, is also provided in Appendix~\ref{sec:app_prompt_engineered}. 
Unless stated otherwise, all results employ the original prompt used for the human data, to mirror the setup.

\paragraph{Data summary.}
Human speakers show clear lexical convergence by round 2, as observed in Figure 2 of \citet{haber-etal-2019-photobook}. To capture this effect without excessive context length, we stopped the VLM self‑play at round 3. From 150 simulated games, we collect 2662 utterances, 101701 tokens. These dialogues constitute the VLM generated corpus used in all subsequent analyses.

\subsection{Extracting Referring Expressions}
\label{ssec:refexp}
Our downstream metrics (§\ref{ssec:alignment}, §\ref{ssec:adaptation}) should operate only on referring expressions, not on meta dialogue (“Ready?”, “Let’s guess”). We therefore processed each utterance to isolate referring expressions. While embedding based filters (BERTScore~\cite{zhang2019bertscore}, and CLIPScore~\citep{hessel-etal-2021-clipscore}) were considered, they are less interpretable. Moreover, we observed that VLMs follow highly regular patterns (\textit{``Image 1 is\ldots''}, \textit{``In my first image\ldots''}), which made rule‑based extraction viable. 

\paragraph{Linking utterances to images.}
In the human corpus, the alignment between each utterance and the image being discussed is obtained from the annotation click logs. Participants indicated which images they considered common or different, which served as a proxy for determining the referent of each referring expression. For the VLM self-play, we explicitly prompted the models to include a \texttt{"reference"} field in their JSON responses (\S~\ref{ssec:dyads}) to indicate which image each utterance pertained to. However, we observed that a substantial proportion of turns included a \texttt{"reference"} field set to \texttt{null}. Nonetheless, most referring expressions were explicit within the utterances themselves which allowed us to leverage the textual content to determine the referent image.

\paragraph{Human validation.}
To validate the pipeline, we randomly sampled 50 rounds (\(\approx10\%\) of the VLM corpus) spanning all three models.  Manual annotation shows 0.99 precision, recall of 0.55, yielding $F1$ score of 0.66. We intentionally prioritized precision in our extraction approach to minimize false positives. This design ensures the integrity of analyses we depend on extracted referring expression, even at the cost of recall.
\section{Metrics}
\label{sec:metrics}
We formalize four families of metrics: \emph{grounding efficiency}, \emph{content alignment}, \emph{lexical adaptation}, and \emph{human-likeness} each grounded in psycholinguistic theory.

\begin{table*}[t]
\centering
\resizebox{.75\linewidth}{!}{
\begin{tabular}{lccc}
\toprule
\textbf{System} & \textbf{Total Score (max 18)} & \textbf{\# Words} & \textbf{\# Turns } \\
\midrule
Claude3.5      & 12.62 $\pm$ 2.07 & 805.72 $\pm$ 123.85 & 15.48 $\pm$ 2.52 \\
GPT4.1     & 15.02 $\pm$ 1.81 & 800.08 $\pm$ 116.87 & 14.68 $\pm$ 2.51 \\
GPT-4o-mini & 13.52 $\pm$ 2.34 & 428.22 $\pm$ 80.74 & 23.08 $\pm$ 2.46 \\
\midrule
Human       & \textbf{16.62} $\pm$ 1.14 & 338.10 $\pm$ 109.37 & 74.08 $\pm$ 12.08 \\
\bottomrule
\end{tabular}
}
\vspace{0.5em}
\caption{Mean $\pm$ standard deviation for total score, number of words, and number of turns per game across systems. Humans achieve the highest task success while using fewer words but significantly more turns. In contrast, VLMs achieve lower task success with longer word counts and fewer turns.}
\label{tab:game_level_results}
\end{table*}

\subsection{Grounding efficiency}
\label{ssec:efficiency}
Psycholinguistic theory suggests that efficient grounding involves refining referring expressions and minimizing unnecessary dialogue over time \citep{clark1991grounding,Brennan}. In human interactions, this is reflected in both reduced lexical effort (fewer words) and more streamlined turn-taking (fewer turns) while maintaining or improving task performance \citep{hawkins-etal-2020-continual}. To evaluate this, we compute:
\begin{itemize}[noitemsep,leftmargin=*,nolistsep]
    \item \textbf{Task success}: Total number of correctly identified common and different images in each round (maximum of 18 points for 3 rounds and 3 images).
    \item \textbf{Word count}: Total number of words produced in each round.
    \item \textbf{Turn count}: Total number of conversational turns in each round.
\end{itemize}

We report grounding efficiency at 1) game level to capture overall communicative cost and task success, and 2) round level dynamics to evaluate how grounding efficiency evolves as interlocutors accumulate shared knowledge over time.

\subsection{Content alignment}
\label{ssec:alignment}
Assessing RQ2, we measure how closely utterances align with the visual referents.

\paragraph{Absolute CLIPScore.}
We compute CLIPScore of the utterance $u$ and the image embedding of the target $img_t$: 
$\textsc{CLIPScore}(u,img_t)$.

\paragraph{Contrastive CLIPScore.}
Psycholinguistic studies show that humans emphasize diagnostic features, which are properties that uniquely identify the target among distractors \citep{dale-1989-cooking, dale1991content, sedivy2003pragmatic}. We capture this with the contrastive score defined as
\begin{align}
  \textsc{ClipCon} =
  \mathrm{CLIPScore}(u,\mathrm{img}_{t}) \notag \\
  - \tfrac{1}{|D|}\!\!\sum_{d\in D}\!\mathrm{CLIPScore}(u,d)
\end{align}
\noindent
where $u$ is the utterance (i.e., referring expression), the divisor $|D|$ converts the raw sum into a mean and makes the score invariant to the number of distractors.

\subsection{Lexical adaptation}
To assess RQ3, we measure whether VLM pairs form human-like conceptual pacts, reusing each other's terms and pruning redundant detail.
\label{ssec:adaptation}

\paragraph{Word Novelty Rate (WNR).}
To quantify how speakers adjust their vocabulary as common ground builds, we adopt the Word Novelty Rate proposed by \citet{hua2024talk}. WNR is a variant of word error rate that counts only insertions and substitutions, and ignores deletions.
Past work shows that interlocutors progressively drop previously established material once it is mutually known \citep{hawkins-etal-2020-continual}. By focusing on insertions and substitutions, WNR captures the moments where a speaker adds or changes wording, i.e.\ where lexical innovation or repair occurs. A declining WNR across rounds indicates that fewer novel words are being introduced, consistent with successful adaptation.

\subsection{Human‑likeness}
\label{ssec:humanlikeness}
\paragraph{Discrete energy distance.} 
To gauge how \emph{human‑like} VLM utterances are at the distributional level, we adopt the Discrete‑Energy Distance of \citet{sicilia-alikhani-2022-leather}. While our previous metrics target specific grounding mechanisms, this distributional measure captures whether the overall distribution of VLM dialogues resemble human interactions. We first embed each game dialogue with
\texttt{all‑MiniLM‑L6‑v2} Sentence Transformer.\footnote{\url{https://huggingface.co/sentence-transformers/all-MiniLM-L6-v2}} This metric compares the average cross group distance (human-VLM) with the average within group distances (human-human, VLM-VLM). Lower energy distance values indicate that the VLM distribution is closer to the human distribution.

\section{Results}
\label{sec:results}
\subsection{Grounding efficiency}
\label{ssec:res_efficiency}
Addressing RQ1, we assess how efficiently VLM pairs establish common ground compared to human speakers. We operationalise grounding efficiency as the balance between communicative cost (measured by total words and turns) and task performance (total score).

\begin{figure*}
    \includegraphics[width=\linewidth]{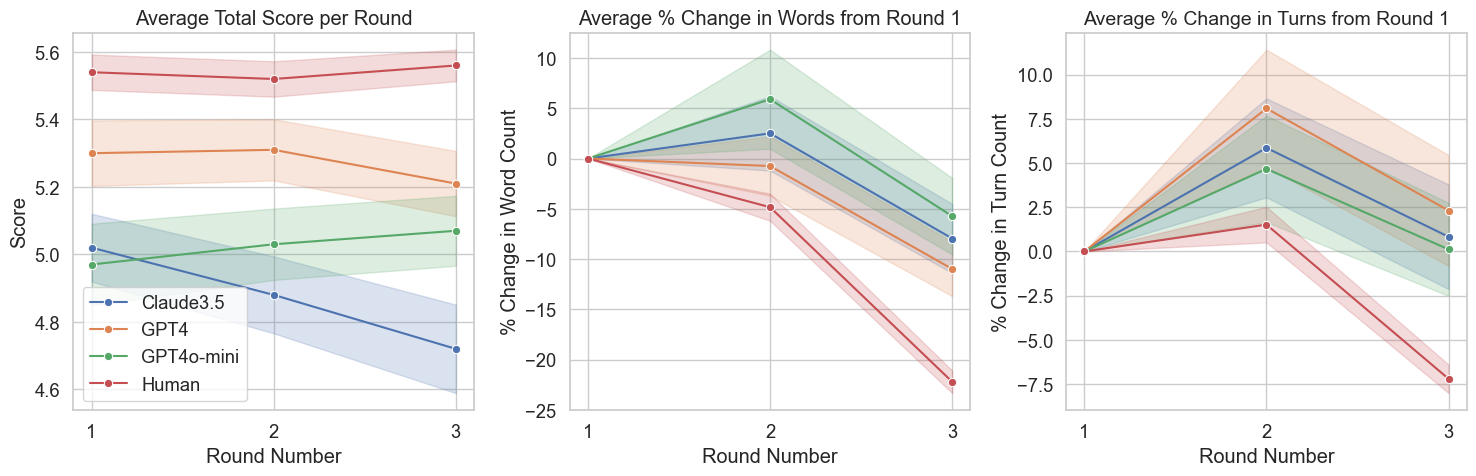}
    \caption{(Left) Average total score per round; (Middle) Average percent change in word count from round 1; (Right) Average percent change in turn count from round 1. Shaded region indicate standard deviation. \textbf{Key takeaways:} (i) Humans and GPT4o-mini improve task performance over rounds; (ii) Humans sharply reduce word count and turn count in later rounds; (iii) VLMs show inconsistent cost reduction, with some increasing word and turn counts.}
    \label{fig:round_results}
\end{figure*}


\paragraph{Game-level performance.}
Table~\ref{tab:game_level_results} summarizes performance across the entire game (capped at 3 rounds) for each system. Humans achieve the highest mean score (16.62), with fewer words (338.1) but more turns (74.08) than VLMs. GPT-4.1 closely approaches human performance in score (15.02) while requiring nearly double the word count and markedly fewer turns. Claude-3.5 shows lower task success (12.62) despite the highest word count (805.72) and reduced turn count.

\paragraph{Round-level performance.}
Figure~\ref{fig:round_results} analyzes grounding efficiency across rounds. We present the average total score per round, percentage change in word count from round 1, and percentage change in turn count from round 1. 

Total score (Fig~\ref{fig:round_results} left) improves with additional rounds for humans and GPT4o-mini. In contrast, GPT4.1 and Claude3.5 exhibit declining scores, possibly due to challenges of managing longer context lengths. Specifically, GPT4.1 and Claude3.5 generate nearly double the total word count per game compared to GPT4o-mini (800 and 806 words vs. 428 words; see Table~\ref{tab:game_level_results}), which may contribute to degraded performance. This observation aligns with prior work demonstrating that LLM performance tends to degrade with longer context windows \cite{an2024does}. 

Word count (Fig~\ref{fig:round_results} middle) consistently decreases across rounds, and declines significantly for humans in round 3. While all models show reduced word count by round 3, GPT4o-mini and Claude3.5 initially increase their word count in round 2, and this pattern contrasts with humans. This is consistent with psycholinguistic theories of lexical entrainment and collaborative efficiency \cite{holler2020communicating}, that the decrease in word count indicates common ground. Further, turn count (Fig~\ref{fig:round_results} right) follows a similar trend that humans generally reduce their turns across rounds, while VLMs tend to increase turn count from round 1 to 2, before exhibiting a slight reduction in round 3. Overall, these patterns suggest that GPT4o-mini demonstrates a grounding trajectory resembling human efficiency with improved task scores with moderate communicative cost, while GPT4.1 and Claude3.5 struggle with verbosity and performance as rounds progress.

\paragraph{Does prompt tuning increase grounding efficiency?}
Motivated by these results, when we specifically ask models to be more human-like in the prompt, we observe that their performance, in fact, becomes closer to humans in various metrics. Specifically, we crafted a revised prompt that preemptively mitigates three recurrent failure modes (\S~\ref{ssec:dyads}), to guide VLMs toward more concise and targeted communication. Results 
(Appendix~\ref{sec:prompt_result}) suggest that tailored prompting can improve efficiency metrics and promote more adaptive behavior similar to humans, though inherent reasoning limitations persist.

\subsection{Content alignment}
\label{ssec:res_alignment}

\begin{figure}[t]
  \centering
  \includegraphics[width=\columnwidth]{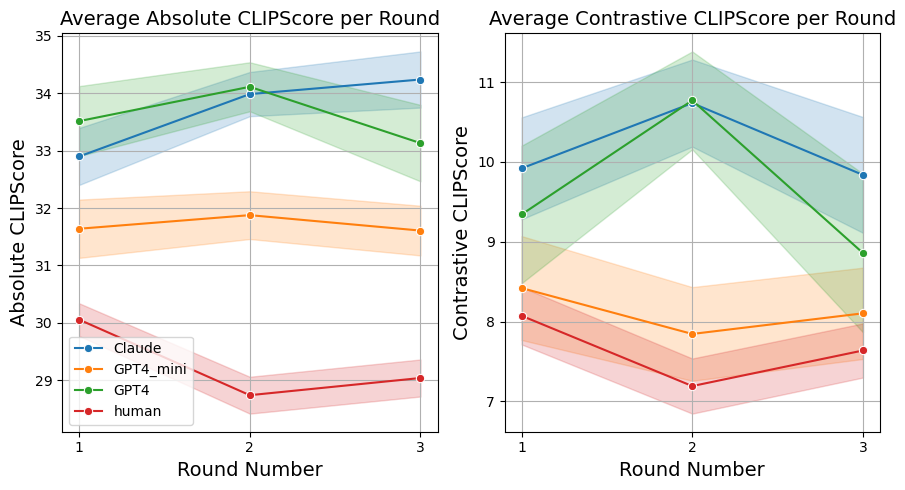}
  \caption{(Left) Absolute CLIPScore. (Right) Contrastive CLIPScore ($\uparrow$ means the utterance is more diagnostic of the target versus distractors).  
  Shaded region indicate standard deviation. \textbf{Takeaway:} Humans steadily lower their CLIPScore while still completing the task, suggesting that they simplify their descriptions as mutual knowledge accrues. LLMs diverge in how they adapt across rounds.}
  \label{fig:clipscore}
\end{figure}

Figure~\ref{fig:clipscore} (left) shows declining absolute CLIPScore for humans. This aligns with prior findings that once common ground is established, speakers economise on explicit visual detail and drop redundant description. GPT4o-mini had a consistent description strategy with absolute CLIPScore of around 31.5 across rounds. In contrast, CLIPScore for Claude3.5 increases across rounds, indicating longer context length might lead the model to provide additional descriptions than prune detail. 

The contrastive variant in Fig~\ref{fig:clipscore} (right) separates the systems further. GPT4.1 and Claude3.5 briefly spike in diagnostic detail during round 2, whereas GPT‑4o‑mini and humans follow a flatter, lower trajectory with lower score in round 2.

\begin{figure}
    \includegraphics[width=\columnwidth]{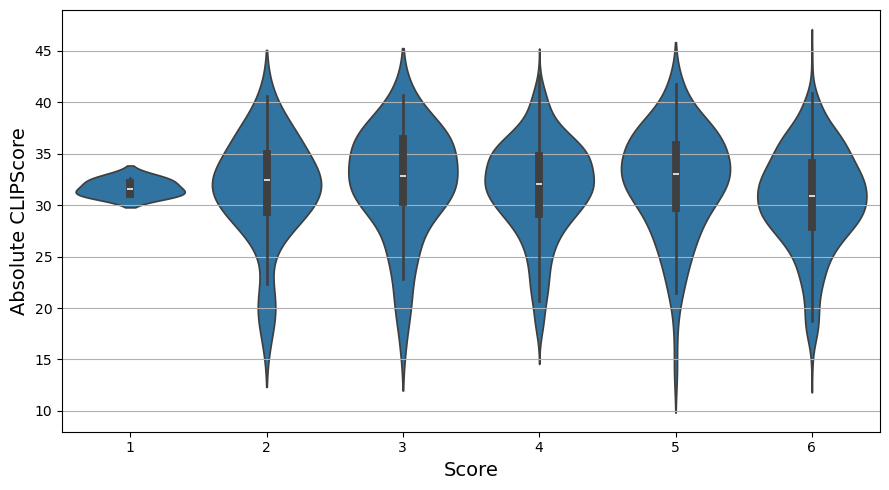}
    \caption{Violin plots show the distribution of Absolute CLIPScore conditioned on the total game score (0–6) in the same round; white horizontal bars mark medians, boxes the inter‑quartile range. \textbf{Takeaway:} High and low alignment scores scattered across all outcome bins confirm that CLIP‑based metrics alone do not predict task success.}
    \label{fig:absolute_clipscore}
\end{figure}

\paragraph{Does alignment drive success?} 
To test whether raw image–utterance similarity translates into better coordination, we relate Absolute CLIPScore to the score actually obtained in the same round (Fig~\ref{fig:absolute_clipscore}). This plot makes the disconnect between alignment and task success explicit, as high and low CLIPScores are scattered across all outcome bins. Results from humans illustrate the point most clearly, as they achieve near‑perfect task scores (Table \ref{tab:game_level_results}) despite the lowest alignment scores (Fig \ref{fig:clipscore}). These observations show that CLIP‑based alignment metrics capture a surface‑level resemblance between words and pixels, but miss the pragmatic reasoning that enables interlocutors to establish common ground.

\subsection{Lexical adaptation}
\label{ssec:res_adaptation}
\paragraph{Word Novelty Rate.} 
\begin{figure}[!h]
    \includegraphics[width=0.95\columnwidth]{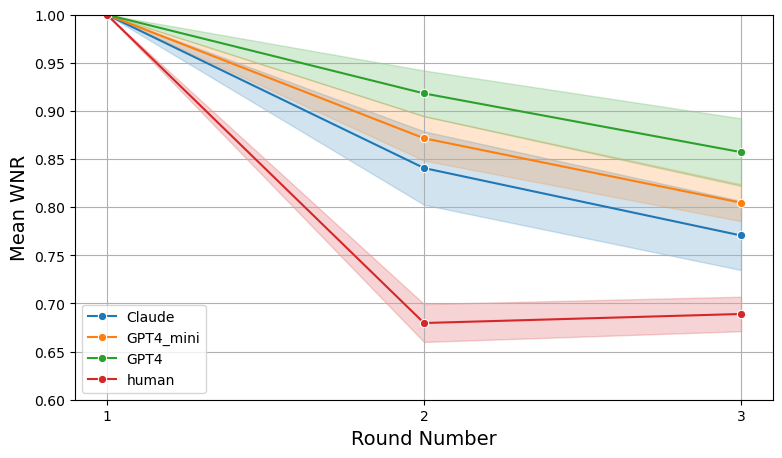}
    \caption{WNR drops over rounds for all systems, but the steepest decline occurs in human referring expressions. Claude and GPT‑4o‑mini show moderate adaptation; GPT‑4 lags behind.}
    \label{fig:wnr}
\end{figure}

Fig~\ref{fig:wnr} shows the mean WNR for referring expressions across rounds. In referring expressions, humans achieve the steepest decline in WNR. This pattern reflects the formation of conceptual pacts and lexical stabilization as common ground builds. Other VLMs show moderate and slower adaptation compared to humans. 

These metrics show that VLM pairs do not fully replicate human strategies of lexical adaptation. While some VLMs, such as Claude, exhibit partial human-like adaptation in lexical choices, GPT-4 models struggle to stabilize and reuse previously grounded referring expressions.

\subsection{Human-likeliness}
\label{ssec:res_humanlikeliness}
\begin{table}[ht]
\centering
\begin{tabular}{lccc}
\hline
\textbf{Model} & \textbf{Energy Distance $\downarrow$} \\
\hline
GPT4.1         & 62\% \\
Claude3.5      & 63\% \\
GPT4-mini    & \textbf{39\%} \\
\hline
\end{tabular}
\caption{Distributional human‑likeness measured by discrete energy distance. Lower values indicate that a model’s utterance distribution is closer to human dialogue. \textbf{Takeaway:} GPT4o‑mini is the most human‑like overall, while Claude 3.5 and GPT‑4.1 remain stylistically farther from human discourse.}
\label{tab:energy_distance}
\end{table}

Table~\ref{tab:energy_distance} lists the energy distance between human dialogues and each VLM pair. GPT4o‑mini attains the smallest distance ($39\%$), which indicates that utterance distribution is the closest to human data. Claude3.5 and GPT4.1 yield substantially higher distances, suggesting that their dialogue style diverges more from human patterns. Although Claude3.5 matched humans in lexical adaptation (\S~\ref{ssec:res_adaptation}), its higher energy distance reveals that its overall dialogue style still diverges from human discourse. 
\begin{figure}
    \centering
    \includegraphics[width=\linewidth]{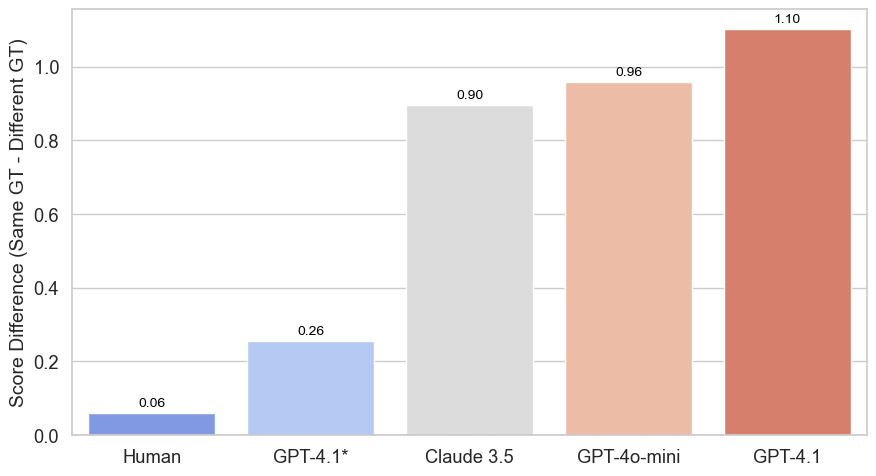}
    \caption{Bar chart showing the performance boost (score difference $\Delta = \text{Same GT} - \text{Different GT}$) for humans and VLMs. Humans exhibit minimal susceptibility ($\Delta=0.06$), while GPT4.1 shows the highest score inflation ($\Delta=1.10$). The prompt-tuned variant GPT4.1* reduces susceptibility ($\Delta=0.26$), suggesting that targeted instructions can mitigate imitation effects.}
    \label{fig:influence}
\end{figure}

\section{Case Study: Sycophantic VLM Guesses}
\label{sec:case_study}
Even though we have used total task score as a proxy for grounding success, we observe that high scores in VLM dialogues can stem from influences from the other interlocutor rather than grounded mutual understanding. Unlike human players, VLMs often exhibit sycophantic behavior, where they adapt their guesses based on their partner’s revealed responses (as illustrated in Figure~\ref{fig:photobook_example}).

Each round requires annotating three images with binary labels (\texttt{C}ommon or \texttt{D}ifferent), yielding only $2^3 = 8$ possible label combinations. It is possible for dyads to coincidentally receive identical ground-truth labels, even if their visual inputs differ. In our sample of 150 rounds, 56 rounds (\textasciitilde 37\%) involved dyads with matching ground-truth labels. If VLMs influence each other’s guesses, such cases can inflate scores and create the illusion of successful grounding.

\paragraph{Score inflation analysis.}
To test whether this was the case, we grouped rounds based on whether the dyad's ground-truth labels were identical or different. We then computed the score difference between these conditions to see the score boost that is attributable to shared ground-truth labels. As shown in Figure~\ref{fig:influence}, humans were robust to shared ground-truth conditions, with only a minor score difference ($\Delta = 0.08$). In contrast, despite achieving the highest average task score among VLMs, GPT4.1 exhibited the greatest susceptibility to this effect, with a score difference exceeding one point ($\Delta = 1.10$). Claude3.5 and GPT4o-mini also exhibited significant susceptibility. 

\paragraph{Mitigation experiment.}
To mitigate this effect, we used a prompt-tuned variant from above (GPT4.1*) that explicitly warned against sharing guesses during the dialogue. This intervention significantly reduced GPT4.1’s susceptibility, to achieve the lowest susceptibility to ground-truth alignment among VLMs. This demonstrates that tailored prompts can mitigate such effects.
\section{Discussion}
\label{sec:discussion}
Our metric suite shows that the current VLMs nearly reproduce the outcomes of human dialogue (as measured with task success), without reproducing the process by which humans achieve those outcomes. Across efficiency (\S\ref{ssec:res_efficiency}), alignment (\S\ref{ssec:res_alignment}), 
 and human-likeliness (\S\ref{ssec:res_humanlikeliness}), GPT4o‑mini consistently approximates human dialogue most closely.  In contrast, while GPT4.1 and Claude3.5 exhibit strengths in task score (\S\ref{ssec:res_efficiency}) and lexical adaptation (\S\ref{ssec:res_adaptation}), respectively, both models exhibit limitations in other metrics.

\paragraph{Why do the models diverge?}
We identify three factors. (i) Training data mismatch. Pre-training corpora contain millions of single image captions but almost no multi-round collaborative dialogues. Consequently, models optimize for listing visual details rather than incremental efforts that characterize human conversation. (ii) Reward alignment bias. RLHF typically rewards "agreeable" or "helpful" completions. When two VLMs converse, this can over penalise informative disagreement and over reward mirroring. Our case study revealed this pattern with inflated task score whenever two VLMs share ground-truth labels (\S\ref{sec:case_study}). (iii) Effortless token generation. For VLMs, generating additional tokens is virtually costless, unlike for humans who face cognitive and temporal constraints. In the absence of incentives for brevity, VLMs tend to produce unnecessarily long utterances and rarely reuse previously established shorthand. This helps explain their limited improvement in grounding efficiency (\S\ref{ssec:efficiency}).

\paragraph{}
\section{Conclusion}
In this paper, we introduced a novel benchmarking approach to assess how effectively VLMs establish common ground through interactive dialogue. Unlike previous evaluations that focus solely on task success, our four-metric suite, grounding efficiency, content alignment, lexical adaptation, and human-likeness, enables a more nuanced examination of VLM performance. Our experiments revealed significant differences between human interactions and VLM self-play, highlighting that achieving high accuracy alone does not imply successful grounding or human-like communicative patterns. 

These findings underscore critical areas for future development, particularly the need for training methods that encourage incremental, collaborative dialogue rather than isolated, verbose responses. Addressing the biases inherent in reinforcement learning alignment methods and incentivizing conciseness could bring VLM interactions closer to human efficiency. By emphasizing the process rather than merely the outcome, our work provides settings for future research aimed at collaborative, human-like AI communication.

\section*{Limitations}
Although our framework broadens the evaluation of VLMs beyond single turn accuracy, several limitations should be noted.

We analyze VLM–VLM dialogues to isolate model capabilities without human guidance. In deployment, systems will converse with humans who provide richer pragmatic cues, error corrections, and social feedback. Whether models adapt differently when paired with human partners remains an open question.

Moreover, because we tested on proprietary VLMs, the underlying architectures, training data, and alignment objectives are opaque. This makes it infeasible to determine whether the observed behaviors arise from model scale, fine-tuning or other design choices. Open source replications with transparent training methods are needed to evaluate the generality of our findings.

\section*{Ethics Statement}
We use the publicly released PhotoBook dataset \citep{haber-etal-2019-photobook}, which contains crowd‑sourced dialogues and MS‑COCO images licensed for research. The dataset does not include personally identifiable information.

All data splits, metric implementations, and analysis scripts will be made publicly available, to enable independent replication and extension.

\section*{Acknowledgments}
This research was supported in part by the U.S. National Science Foundation under Award No. 2418664. We thank Asteria Kaeberlein for her helpful feedback.

\bibliographystyle{acl_natbib}

\appendix
\section{Prompts}
\subsection{Original prompt}
\label{sec:app_prompt}
You will have a conversation with another player to determine which of the six images on your display are shared between both of you, and which are unique to your list.  
\textbf{Objective:} Identify the common and different images as quickly as possible!

\vspace{1ex}

\noindent Your images:
\begin{itemize}
    \item \textbf{Image 1}: \texttt{image\_filename.jpg}
    \item \textbf{Image 2}: \texttt{image\_filename.jpg}
    \item \textbf{Image 3}: \texttt{image\_filename.jpg}
\end{itemize}

\vspace{1ex}

\noindent\textbf{Instructions:} Return a JSON field \texttt{"guesses"} containing exactly three letters:
\begin{verbatim}
"guesses": ["C", "D", "C"]
\end{verbatim}
where each position corresponds to:
\begin{itemize}
    \item \textbf{Image 1}
    \item \textbf{Image 2}
    \item \textbf{Image 3}
\end{itemize}

\noindent with the following meanings:
\begin{itemize}
    \item \textbf{C}: The picture is present in both your list \textit{and} your partner’s list.
    \item \textbf{D}: The picture appears only in your list.
\end{itemize}

\vspace{1ex}

\noindent\textbf{Output format}: Every turn must consist of a single valid JSON object containing:
\begin{itemize}
    \item \texttt{"message"} – your utterance as plain text
    \item \texttt{"reference"} – \texttt{"Image 1"}, \texttt{"Image 2"}, \texttt{"Image 3"}, or \texttt{null}
    \item \texttt{"guesses"} – either \texttt{null} \textbf{or} a list like \texttt{["C", "D", "C"]}
\end{itemize}

\subsection{Engineered prompt}
\label{sec:app_prompt_engineered}
Goal: find which images are \textbf{COMMON} between you and your partner's list.  
You are \textbf{Player A}.  
Refer to images \textit{only} as ``Image~1'', ``Image~2'', or ``Image~3'' -- never mention file names or IDs.  
\textbf{Important:} the number (Image~1/2/3) is \textit{local} to you. Your Image~1 could be your partner’s Image~3.  
Decide only from what you SEE, not from the number.  
\quad -- Describe what you see in each image you are currently discussing.  
\quad -- Ask short, precise questions about your partner’s view.  
\quad -- Answer their questions clearly.

\vspace{1ex}

\noindent Your images:  
\begin{itemize}
    \item \textbf{Image 1}: \texttt{image\_filename.jpg}
    \item \textbf{Image 2}: \texttt{image\_filename.jpg}
    \item \textbf{Image 3}: \texttt{image\_filename.jpg}
\end{itemize}

\vspace{1ex}

\noindent When -- and \textit{only} when -- you are \textbf{confident about \textit{all three} images},  
return a JSON field \texttt{"guesses"} containing exactly three letters:
\begin{verbatim}
"guesses": ["C", "D", "C"]
\end{verbatim}
where each position corresponds to \textbf{Image~1~Image~2~Image~3}  
in that order:  
\quad $\bullet$ \textbf{C} = the picture is present in both your list \textit{and} your partner’s list  
\quad $\bullet$ \textbf{D} = the picture appears only in your list

\vspace{1ex}

\noindent The physical position of the matching image in your partner’s list does \textbf{not} matter.  
Once you send a non-null \texttt{guesses}, you may not revise them.  
Sending a final \texttt{guesses} automatically signals that you are \textit{ready} to move on.  
\textbf{Do NOT reveal or hint at your provisional guesses.}  
Accuracy first; shorter dialogues and fewer turns earn higher scores.

\vspace{1ex}

\noindent \textbf{Output format} -- every turn must be a single valid JSON object with:
\begin{itemize}
    \item \texttt{"message"} -- your utterance as plain text
    \item \texttt{"reference"} -- ``Image 1'', ``Image 2'', ``Image 3'', or \texttt{null}
    \item \texttt{"guesses"} -- \texttt{null} \textbf{or} a list like \texttt{["C","D","C"]}
\end{itemize}

\section{Prompt Engineering Results}
\label{sec:prompt_result}
\begin{figure*}
    \centering
    \includegraphics[width=\linewidth]{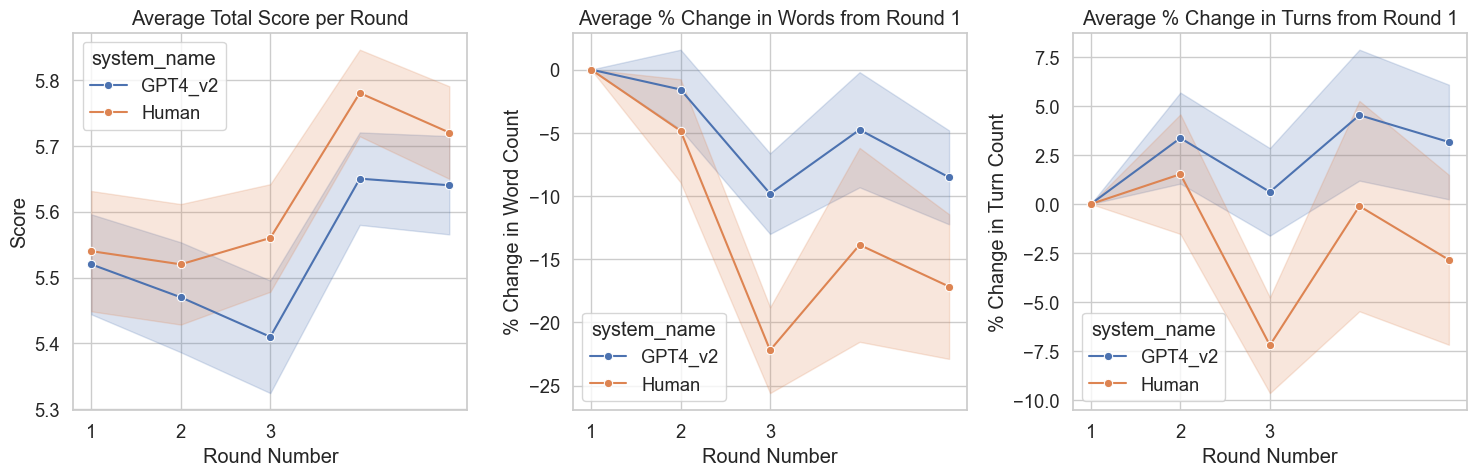}
    \caption{Comparative performance of GPT4.1* and humans. 
    \textbf{Left:} Average total score per round, indicating gradual improvement for both systems, with humans consistently achieving higher scores. 
    \textbf{Middle:} Average percent change in word count from Round 1. The plot shows humans' sharper reduction in word use compared to GPT4.1*, though the model shows a similar declining trend. 
    \textbf{Right:} Average percent change in turn count from Round 1, where humans sharply decrease the number of conversational turns needed as rounds progress. GPT4.1* follows a similar adaptive trajectory with the revised prompt, but with a more gradual reduction.}
    \label{fig:prompt_engineering}  
\end{figure*}
Figure~\ref{fig:prompt_engineering} compares the performance of GPT4.1* (with revised prompting) and humans. The analysis shows that while humans naturally achieve higher scores, greater reductions in word count, and fewer conversational turns as rounds progress GPT4.1* when guided by targeted prompts begins to mirror these adaptive trends. Although the model’s improvements are less steep, its gradual reduction in word use and turns suggests that prompt engineering can steer VLMs towards more efficient, human-like grounding behavior. These results highlight that while inherent model limitations persist, careful prompt design can significantly enhance adaptive communication patterns.

\section{Additional Lexical Adaptation Metric}
\subsection{Kullback-Leibler Divergence.}
Let $P_{g,r}(w)$ be the unigram distribution of referring expression tokens in round $r$, $P_{g,1}(w)$ its counterpart in the first round.  Lexical convergence is measured by
$$
  D_{\mathrm{KL}}(P_{g,1}\,\Vert\,P_{g,r})
  = \sum_{w} P_{g,1}(w)\,
    \log\frac{P_{g,1}(w)}{P_{g,r}(w)}
$$
as introduced by \citep{kullback1951information}. This divergence quantifies how much the token distribution in a later round departs from the initial distribution. A low $D_{\mathrm{KL}}$ value indicates convergence, that speakers reuse terms and maintain a shared vocabulary established early in the game. Our analysis computes these distributions directly from tokenized dialogue, and apply Laplace smoothing ($\epsilon=10^{-8}$) to handle rare or unseen tokens. For each model, we compute these KL values over all games and report the mean and standard error, and compare with human behavior.
\subsection{Results.}
\begin{figure}
    \includegraphics[width=\columnwidth]{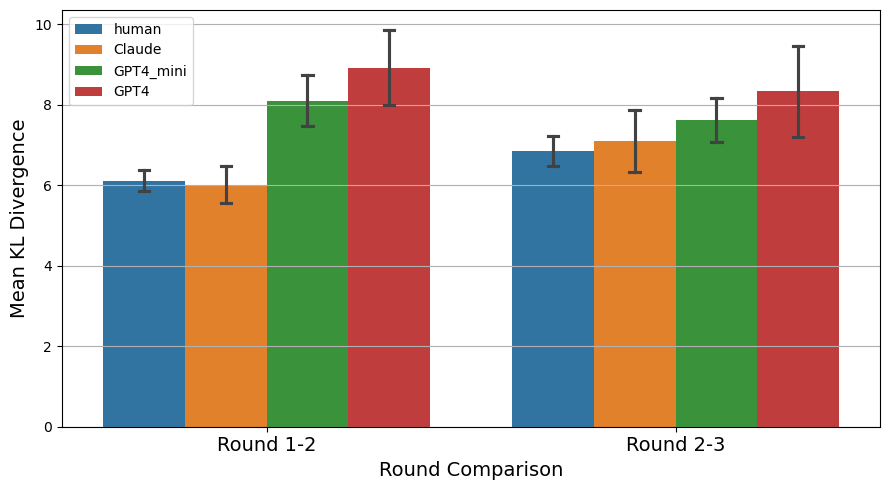}
    \caption{Bars show mean divergence for round 1–2 and 2–3 comparisons; error bars denote standard error. Lower values indicate stronger lexical convergence. Humans and Claude show consistent adaptation; GPT‑4 models converge more slowly.}
    \label{fig:kl_divergence}
\end{figure}

\paragraph{KL divergence between consecutive rounds.} Figure~\ref{fig:kl_divergence} shows the mean KL divergence between rounds 1–2 and 2–3 for each system, computed over referring expressions. Claude’s divergence is similarly low as humans for both comparisons. This suggests that it can also align its referring expressions across rounds. Moreover, while GPT4o-mini and GPT4.1 show higher divergence in round 1-2, they converge slightly between round 2-3. 

\end{document}